\documentclass[journal]{IEEEtran}

\usepackage{graphicx}
\usepackage{cite}
\usepackage{picinpar}
\usepackage{amsmath}
\usepackage{url}
\usepackage{flushend}
\usepackage[utf8]{inputenc}
\usepackage{colortbl}
\usepackage{soul}
\usepackage{multirow}
\usepackage{pifont}
\usepackage{color}
\usepackage{alltt}
\usepackage[hidelinks]{hyperref}
\usepackage{enumerate}
\usepackage{siunitx}
\usepackage{breakurl}
\usepackage{epstopdf}
\usepackage{pbox}
\usepackage{subfigure}
\usepackage{mathrsfs,amsmath}
\usepackage{bm}
\usepackage{algorithm,float,algorithmic}
\usepackage[english]{babel}
\usepackage{amsthm}
\usepackage{amssymb}
\usepackage{caption}
\usepackage{xspace}
\usepackage{threeparttable}
\usepackage{booktabs}

\begin{document}
\title{LoRM: Learning the Language of Rotating Machinery for Self-Supervised Condition Monitoring}%
\author{
Xiao~Qin,
Xingyi~Song,
Tong~Liu,
Hatim~Laalej,
Zepeng~Liu\textsuperscript{*},
Yunpeng~Zhu\textsuperscript{*},
and Ligang~He\textsuperscript{*}%
\thanks{\textsuperscript{*}Corresponding authors: Zepeng Liu (e-mail: zepengliu@tongji.edu.cn), Yunpeng Zhu (e-mail: yunpeng.zhu@qmul.ac.uk), and Ligang He (e-mail: ligang.he@warwick.ac.uk).}%
\thanks{Xiao Qin and Ligang He are with the Department of Computer Science, University of Warwick, Coventry, UK (e-mail: xiao.qin@warwick.ac.uk; ligang.he@warwick.ac.uk).}%
\thanks{Yunpeng Zhu is with the School of Engineering, Queen Mary University of London, London, UK (e-mail: yunpeng.zhu@qmul.ac.uk).}%
\thanks{Xingyi Song and Tong Liu are with the Department of Computer Science, University of Sheffield, Sheffield, UK (e-mail: x.song@sheffield.ac.uk; T.Liu@sheffield.ac.uk).}%
\thanks{Hatim Laalej is with the Advanced Manufacturing Research Centre, University of Sheffield, Sheffield, UK (e-mail: h.laalej@sheffield.ac.uk).}%
\thanks{Zepeng Liu is with the Shanghai Research Institute for Intelligent Autonomous Systems, State Key Laboratory of Autonomous Intelligent Unmanned Systems, Department of Control Science and Engineering, Tongji University, Shanghai, China (e-mail: zepengliu@tongji.edu.cn).}%
}

\maketitle
    
\begin{abstract}
We present LoRM (Language of Rotating Machinery), a self-supervised framework for multi-modal rotating-machinery signal understanding and real-time condition monitoring. LoRM is built on the idea that rotating-machinery signals can be viewed as a machine language: local signals can be tokenised into discrete symbolic units, and their future evolution can be predicted from observed multi-sensor context. Unlike conventional signal-processing methods that rely on hand-crafted transforms and features, LoRM reformulates multi-modal sensor data as a token-based sequence-prediction problem. For each data window, the observed context segment is retained in continuous form, while the future target segment of each sensing channel is quantised into a discrete token. Then, efficient knowledge transfer is achieved by partially fine-tuning a general-purpose pre-trained language model on industrial signals, avoiding the need to train a large model from scratch. Finally, condition monitoring is performed by tracking token-prediction errors as a health indicator, where increasing errors indicate degradation. In-situ tool condition monitoring (TCM) experiments demonstrate stable real-time tracking and strong cross-tool generalisation, showing that LoRM provides a practical bridge between language modelling and industrial signal analysis. The source code is publicly available at \url{https://github.com/Q159753258/LormPHM}.
\end{abstract}


\section{Introduction}
\IEEEPARstart{R}{otating} machinery is fundamental to manufacturing, energy, transportation, and process industries, where reliability and operational safety are of vital importance \cite{Peng_Automatic_Feature_Ex}. Unplanned downtime or catastrophic failures can cause substantial economic losses and safety risks, including material waste, production interruption, and severe accidents \cite{Zhang_Integrated_Multitasking}. Therefore, timely and accurate machine health assessment through real-time condition monitoring is essential to prevent minor degradation from developing into serious system failures \cite{Zhao_Lifelong_Monitoring}.

In practical industrial environments, rotating machinery such as motors, turbines, gearboxes, and machine tools often operate under dynamic loads, strong disturbances, and complex nonlinear conditions. Under such circumstances, a single sensor is often insufficient to fully characterise machine behavior because of measurement noise, nonlinear responses, and environmental interference \cite{Wang_Review_Sensor}. As a result, multi-modal sensing has become an effective strategy, where vibration, acoustic, force, electrical, and other measurements are jointly used to provide a more comprehensive description of machine behavior \cite{Matthew_Unsupervised_multimodal}. However, these heterogeneous signals differ in physical meaning, dynamic range, bandwidth, and noise characteristics \cite{Ruan_Survey_Multisensor}. Therefore, the effective integration of heterogeneous sensor signals for reliable and accurate real-time condition monitoring remains a major challenge for both academia and industry.

\begin{figure}[t]
  \centering
  \includegraphics[width=0.98\columnwidth]{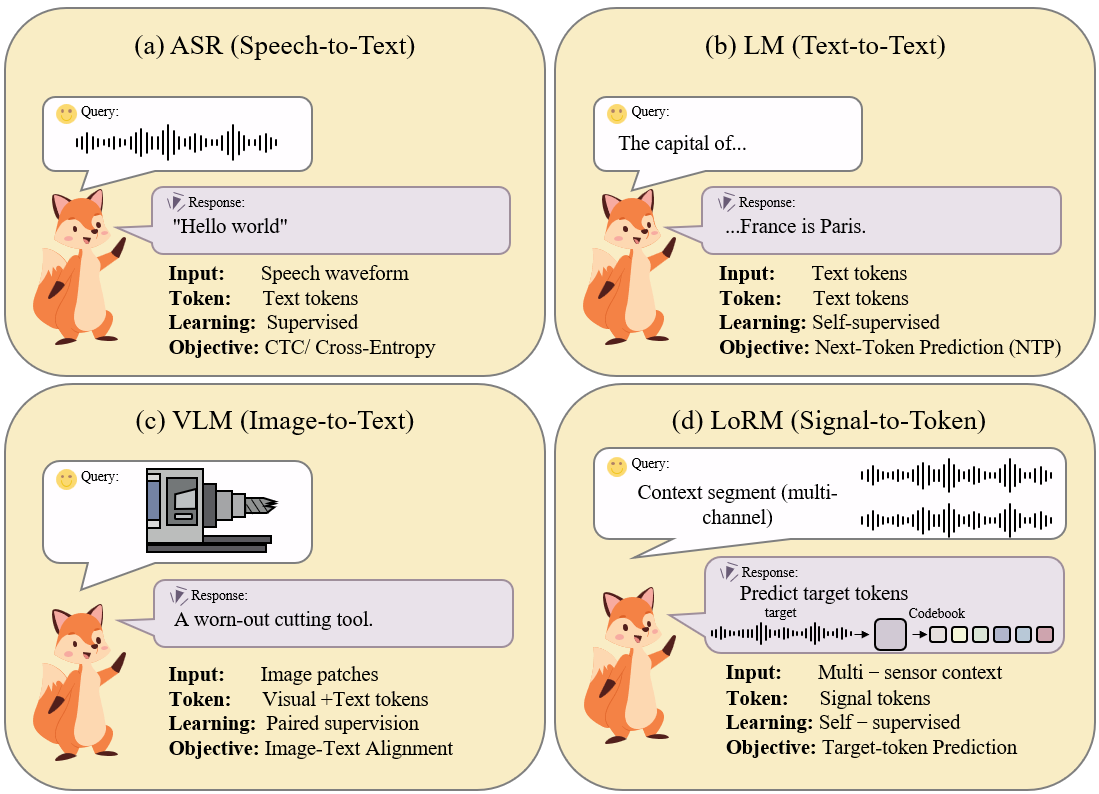}
  \caption{Training through token prediction: from ASR, LMs, and VLMs to LoRM. LoRM extends this paradigm to rotating machinery by predicting discrete future signal tokens from observed multi-sensor context.}
  \label{fig:Token Prediction}
\end{figure}

To address this challenge, multi-sensor fusion techniques have been extensively studied at the data, feature, and decision levels \cite{Lin_Review_Multi_Sensor}. Data-level fusion directly combines raw signals or transformed spectra \cite{Pan_Fast_fault,Azamfar_Multisensor_fusion,Cheng_Fusion_method}, but usually requires strict temporal alignment and high computational cost. Feature-level fusion relies on time-domain, frequency-domain, or time-frequency features \cite{Liu_Sensor_data,Gui_LRLASLA}, but often depends heavily on feature quality and expert knowledge. Decision-level fusion aggregates the outputs of independent classifiers \cite{Tong_Ensemble_Learning}, but may neglect cross-modal dependencies and propagate errors from weak modalities. As a result, most existing approaches still struggle to learn unified representations of multi-sensor machine dynamics. Recent progress in foundation models suggests a useful direction for this problem. As illustrated in Fig.~\ref{fig:Token Prediction}, many modern models across different modalities follow a similar paradigm: raw inputs are transformed into modelled units, such as tokens or patches, and contextual dependencies are learned through self-supervised prediction. Recent studies further show that this paradigm can also be extended to time-series modelling with pre-trained Transformer backbones \cite{Brown_Language_models,Bert,BART,Naveed_overview_LLM,Chronos,Moment}.

However, most existing time-series foundation models are designed primarily for single-modality and relatively clean structured datasets. In contrast, industrial signals from rotating machinery are inherently multi-modal, high-frequency, and noise-rich, which makes direct adaptation of existing models highly challenging. More importantly, current studies rarely explain why industrial signals should be formulated in a language-like manner or what specific properties make such a formulation meaningful.

Motivated by these observations, we propose LoRM (Language of Rotating Machinery), a self-supervised framework that extends the token-prediction paradigm to rotating machinery signals. As shown in Fig.~\ref{fig:Token Prediction}(d), LoRM preserves observed multi-modal context segments in continuous form and predicts discrete future signal tokens. This formulation is supported by several language-like properties of rotating machinery signals. 
\begin{enumerate}
    \item Continuous sensor signals often contain recurring local patterns that can be grouped into a finite number of representative types and arranged as temporal sequences \cite{Antoni_Cyclostationarity}.
    \item These pattern units are not independent, because the recent signal history constrains the near-future evolution through the underlying machine dynamics \cite{Randall_bearing_tutorial}. 
    \item Different sensing channels capture different aspects of the same physical process and therefore provide complementary information for contextual modelling. 
\end{enumerate}
 Therefore, these properties suggest that multi-modal machine signals can be reformulated as a structured token-prediction problem conditioned on observed context. 

In LoRM, each sensor data window is divided into two parts: a context segment, which preserves the observed continuous waveform and provides recent temporal context, and a target segment, which contains the subsequent signal pattern to be discretised and predicted. Following the speech-to-text principle \cite{Speech_to_text}, LoRM converts the target segment of each sensing channel into a symbolic token and trains the model to predict these future tokens from the multi-sensor context segment. In addition, inspired by code-switching in multilingual communication \cite{Code_switching}, different sensing channels, such as vibration, acoustic, and force, are treated as complementary ``modality languages''. Their context segments are flattened in a predefined order to form a unified input sequence, allowing a shared Transformer backbone to jointly model temporal and cross-channel dependencies. A lightweight prediction head is attached to each modality to estimate its target token, and the overall training objective aggregates token-prediction losses across channels. Under healthy operating conditions, future target-segment tokens follow relatively stable transition patterns conditioned on the preceding context segment and can therefore be predicted with low error. As degradation develops, the collected signals gradually deviate from the normal dynamics learned by the model, making target-token prediction increasingly difficult. The resulting increase in prediction error can thus serve as an effective indicator of abnormal behavior, enabling self-supervised condition monitoring without requiring dense fault targets.

The main contributions of this work are summarised as follows.
\begin{enumerate}
	\item A new concept, termed the language of rotating machinery, is proposed to interpret multi-sensor industrial signals as tokenisable and context-dependent dynamic sequences, thereby providing a unified language-modelling view for self-supervised condition monitoring.
	\item A self-supervised LoRM framework is developed, in which each signal window is divided into a continuous context segment and a discrete target segment, enabling future machine-state token prediction from observed multi-sensor context.
	\item A code-switching-inspired multi-sensor fusion mechanism is introduced, where different sensing channels are treated as complementary modality languages and combined into a shared contextual sequence for unified temporal and cross-channel modelling.
    \item A transfer-learning strategy is adopted to adapt PLMs to industrial multi-sensor data efficiently, avoiding the cost of training large models from scratch.
	\item In-situ industrial case studies verify that the proposed LoRM framework achieves accurate and generalisable self-supervised condition monitoring.
\end{enumerate}

\section{Basic Idea: The Language of Rotating Machinery}

\subsection{Concept}
Signals collected from rotating machinery often exhibit periodic or cyclo-stationary patterns, meaning that the recent past typically contains useful information for predicting the near future \cite{Antoni_Cyclostationarity,Randall_bearing_tutorial}. Building on this characteristic, LoRM interprets multi-sensor signals, such as vibration, acoustic, electrical, and force measurements, as a learnable language describing machine behavior. Each fixed-length signal window is divided into two parts: a \emph{context segment}, which retains continuous contextual information, and a \emph{target segment}, which is discretised into symbolic tokens. The language model is then trained to predict the target tokens from the context segment, in a manner analogous to speech-to-text systems that convert continuous audio waveforms into discrete linguistic units.

\subsection{Context--target Formulation}
Consider a multi-sensor signal window $\mathbf{X} = [x_{t}^{(c)}]_{t=1,\ldots,W;\; c=1,\ldots,C} \in \mathbb{R}^{W \times C}$, where $W$ is the window length and $C$ is the number of sensor channels. The element $x_{t}^{(c)}$ denotes the measurement of channel $c$ at time index $t$.

A split index $S$ divides the window into a context segment $\mathbf{X}_{\text{ctx}} \in \mathbb{R}^{S \times C}$, which remains in continuous form and provides recent contextual information, and a target segment $\mathbf{X}_{\text{tgt}} \in \mathbb{R}^{(W-S) \times C}$ containing subsequent samples. For each channel $c$, the target waveform is quantised into a single discrete token $y^{(c)} \in \{1,\ldots,K\}$ by applying $k$-means clustering to construct a channel-specific codebook, where $K$ is the number of pattern types for that channel. The target segment of a multi-sensor window is thus represented by the token vector $\mathbf{y} = (y^{(1)},\ldots,y^{(C)})$.

The model is trained to predict each channel token from the continuous context segment, that is, to estimate the conditional probability $p(y^{(c)} \mid \mathbf{X}_{\text{ctx}})$, which links recent signal behavior to the most likely future state of each channel.

For multi-sensor fusion, the continuous context segments from all channels are flattened into a single combined sequence and processed by a shared pre-trained backbone. A lightweight prediction head outputs $p(y^{(c)} \mid \mathbf{X}_{\text{ctx}})$ for each channel, and the overall loss aggregates the token-prediction errors across all channels.

\subsection{Why ``language''?}

In LoRM, the term ``language of rotating machinery'' denotes a modelling formulation shown in Table \ref{lorm_language_analogy}. As can be seen, discrete target tokens play the role of words, a finite per-channel codebook forms the vocabulary, and the multi-sensor context segment provides the contextual information from which future states are inferred.

This formulation is also motivated by practical considerations. The raw rotating machinery signals are typically noisy, and due to the dynamics of rotating machinery, many waveform patterns exhibit recurring structures. These patterns can therefore be represented by a finite set of "Tokens". Tokenisation provides a form of mapping continuous waveform segments into a discrete codebook, while reducing sensitivity to noise and amplitude variations. 

Under this formulation, the learning objective is target-token prediction, which is consistent with the core mechanism of language modelling. When implemented with a Transformer backbone, self-attention enables the model to learn interactions across time and sensing channels within the unified sequence, while a longer context segment provides broader historical information for capturing longer-range dependencies. In this sense, ``language'' refers to tokenised future signal patterns predicted from multi-sensor context, rather than to linguistic semantics.

\begin{table}[t]
\caption{Analogy between language-modelling concepts and LoRM.}
\label{lorm_language_analogy}
\centering
\begin{tabular}{p{0.4\linewidth} p{0.5\linewidth}}
\hline
\textbf{Language modelling} & \textbf{LoRM for rotating machinery} \\
\hline
Word / token & Discretised target token $y^{(c)} \in \{1,\ldots,K\}$ for channel $c$ \\
Vocabulary & Per-channel codebook of $K$ clusters learned by $k$-means \\
Context & Continuous multi-sensor context segment $\mathbf{X}_{\text{ctx}}$ \\
Token prediction & Predict $p(y^{(c)} \mid \mathbf{X}_{\text{ctx}})$ for each channel \\
Sequence (sentence) & Flattened context sequence from all channels (multi-sensor context) \\
Self-attention & Models interactions across time and channels within the unified sequence \\
Long-range dependency & Longer context segment provides broader history \\
\hline
\end{tabular}
\end{table}

\begin{figure*}[t]
  \centering
  \includegraphics[width=0.9\linewidth]{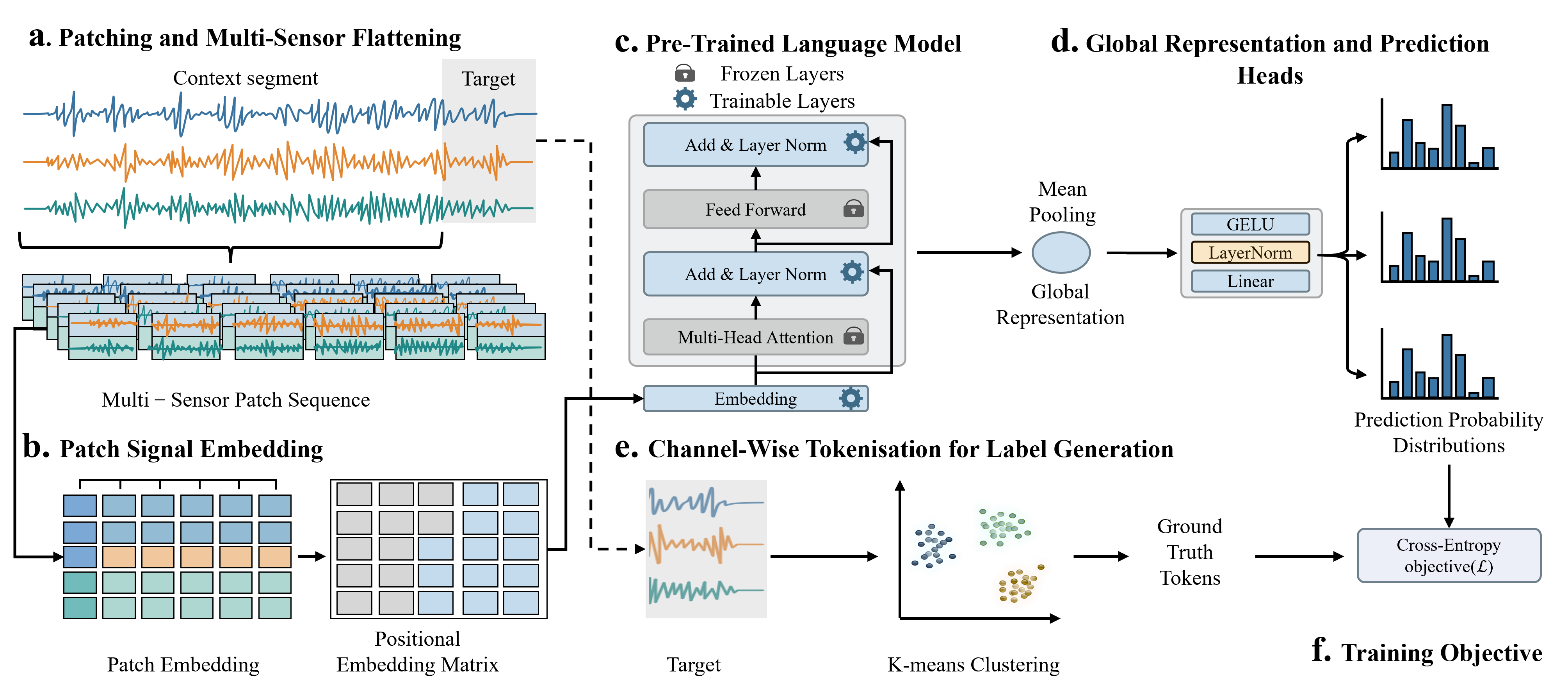}
  \caption{Overall architecture and learning mechanism of LoRM. (a) A multi-channel signal window is divided into a context segment and a target segment, and the context segment is partitioned and flattened into a multi-sensor patch sequence. (b) The patch sequence is projected into the hidden space of a pre-trained language model with positional embeddings. (c) A partially fine-tuned pre-trained language model encodes temporal and cross-channel dependencies. (d) The contextualised patch features are mean-pooled into a global representation and transformed into channel-wise token probability distributions. (e) The target segment of each channel is independently tokenised by $k$-means clustering to generate ground-truth tokens. (f) A cross-entropy objective is applied between the predicted distributions and the generated tokens for self-supervised training.}
  \label{fig:online_training_flow}
\end{figure*}

\section{LoRM Architecture and Learning Mechanism}
\label{sec:method}
This section outlines the LoRM framework, including the construction of the multi-sensor patch sequence, patch embedding, PLM-based contextual modelling, channel-wise prediction heads, and the overall learning mechanism.

\subsection{Patching and Multi-Sensor Flattening}
\label{subsec:patching}
For channel $c$, the context segment of length $S$ is divided along the time axis into $N=\lfloor S/h \rfloor$ non-overlapping patches of length $h$:
\begin{equation}
\mathbf{p}_{n}^{(c)}
=
\big[ x_{t}^{(c)} \big]_{t=(n-1)h+1:nh}
\in \mathbb{R}^{h\times 1},
\qquad n=1,\ldots,N
\label{eq:patch_def}
\end{equation}
where zero padding is applied when necessary.

After patching all $C$ channels, the patches are arranged in a predefined channel-major order and stacked into a unified sequence:
\begin{equation}
\begin{aligned}
\mathbf{P}
=&\;
\big[
\mathbf{p}_{1}^{(1)}, \ldots, \mathbf{p}_{N}^{(1)},
\mathbf{p}_{1}^{(2)}, \ldots, \mathbf{p}_{N}^{(2)}, \ldots, \\
&\;
\mathbf{p}_{1}^{(C)}, \ldots, \mathbf{p}_{N}^{(C)}
\big]^{\top}
\in \mathbb{R}^{NC\times h}
\label{eq:patch_seq}
\end{aligned}
\end{equation}
where $\mathbf{P}$ denotes the multi-sensor context patch sequence (MCPS). This flattening operation is conceptually related to code-switching, where different sensing modalities are interleaved within a shared sequence for joint contextual modelling. Patching also reduces the effective sequence length from $S$ to $N=S/h$, thereby lowering the self-attention cost from $O(S^2)$ to $O(S^2/h^2)$.



\subsection{Patch Signal Embedding}
\label{subsec:embedding}
To make the patch sequence compatible with the PLM input format, a learnable linear embedding layer is used to project each patch into the $d$-dimensional hidden space:
\begin{equation}
\label{eq:embed}
\mathbf{E} = \mathbf{P}\mathbf{W}_{E},
\qquad
\mathbf{W}_{E} \in \mathbb{R}^{h \times d},\;
\mathbf{E} \in \mathbb{R}^{NC \times d}
\end{equation}
where $\mathbf{P}$ is the MCPS, $\mathbf{W}_{E}$ is the embedding matrix, and $\mathbf{E}$ is the embedded patch matrix (EPM). This step converts raw patch signals into fixed-length vectors for Transformer-based contextual modelling.

\subsection{Pre-Trained Language Model Adaptation}
\label{subsec:plm}

After obtaining the EPM, LoRM utilises a pre-trained language model (PLM) to capture dependencies across patches. In this paper, PLM refers to a Transformer-based foundation model pre-trained on large-scale general-purpose corpora (text or speech). A positional embedding $\mathbf{P}_{\mathrm{pos}}$ is added to $\mathbf{E}$ to encode the temporal order and channel identity of patches before it is processed by the PLM:
\begin{equation}
\label{eq:add_pos}
\tilde{\mathbf{E}}
=
\mathbf{E} + \mathbf{P}_{\mathrm{pos}},
\qquad
\mathbf{P}_{\mathrm{pos}} \in \mathbb{R}^{NC \times d}
\end{equation}
where $\tilde{\mathbf{E}}$ denotes the position-enhanced matrix, which is then processed by the PLM, yielding the contextualised patch matrix $\mathbf{Z}$:
\begin{equation}
\label{eq:plm}
\mathbf{Z}
=
\mathrm{PLM}(\tilde{\mathbf{E}}),
\qquad
\mathbf{Z} \in \mathbb{R}^{NC \times d}
\end{equation}
where each row of $\mathbf{Z}$ summarises the contextual information of the corresponding patch.

Following the partial-freezing strategy used in GPT4TS \cite{zhou2023onefitsall}, LoRM adopts partial fine-tuning: the attention and feed-forward layers of the PLM remain frozen, while only the embedding matrix $\mathbf{W}_{E}$, positional embedding $\mathbf{P}_{\mathrm{pos}}$, and normalisation layers are trainable. This reduces the number of trainable parameters and improves training efficiency.

\subsection{Global Representation and Prediction Heads}
\label{subsec:heads}

Given the contextualised patch matrix $\mathbf{Z}$, the global context representation $\mathbf{g}$ is computed by averaging all patch features:
\begin{equation}
\label{eq:pool}
\mathbf{g}
=
\frac{1}{NC}\,\mathbf{1}^{\top}\mathbf{Z} \in \mathbb{R}^{1 \times d}
\end{equation}
where $\mathbf{1} \in \mathbb{R}^{NC \times 1}$ is an all-ones vector. The vector $\mathbf{g}$ summarises the contextual information from all patches.

A lightweight prediction head is applied to the global representation $\mathbf{g}$ to produce token scores for all channels. The head consists of a GELU activation, Layer Normalisation, and a linear classifier:
\begin{equation}
\label{eq:head}
\mathbf{u}
=
\mathrm{LayerNorm}(\mathrm{GELU}(\mathbf{g})),
\qquad
\mathbf{v}
=
\mathbf{u}\,\mathbf{W}_{c}
\end{equation}
where $\mathbf{W}_{c} \in \mathbb{R}^{d \times KC}$ is the classification matrix shared across channels, $\mathbf{u} \in \mathbb{R}^{1 \times d}$ is the intermediate feature, and $\mathbf{v} \in \mathbb{R}^{1 \times KC}$ is the resulting unnormalised score vector. The scores in $\mathbf{v}$ are arranged as $C$ consecutive blocks, with each block containing the $K$ token scores for one sensor channel.

For channel $c$, the predicted token probability distribution is obtained by normalising its corresponding score block with the softmax operator:
\begin{equation}
\label{eq:softmax}
\pi^{(c)}
=
\mathrm{softmax}\!\left(
\mathbf{v}_{(c-1)K+1 :\, cK}
\right)
\in \mathbb{R}^{K}
\end{equation}
where $\pi^{(c)}$ denotes the predicted probability distribution over the $K$ tokens for channel $c$.

As a result, LoRM uses the shared global representation $\mathbf{g}$ to model cross-channel dependencies through the PLM backbone, while producing channel-wise token predictions through separate score blocks.

\subsection{Channel-Wise Tokenisation for Label Generation}
\label{subsec:tokenisation}

To achieve self-supervised learning, the target segment $\mathbf{X}_{\text{tgt}} \in \mathbb{R}^{(W-S) \times C}$ of each window is converted into discrete tokens. Since different sensor channels exhibit distinct physical characteristics and operate over different value ranges, tokenisation is performed independently for each channel.

For channel $c$, all target segments in the training set are collected and clustered using the $k$-means algorithm with $K$ clusters to construct a channel-specific codebook:
\begin{equation}
\mathcal{C}^{(c)}
=
\{
\boldsymbol{\mu}^{(c)}_{1},
\boldsymbol{\mu}^{(c)}_{2},
\ldots,
\boldsymbol{\mu}^{(c)}_{K}
\}
\end{equation}
where $\boldsymbol{\mu}^{(c)}_{k} \in \mathbb{R}^{(W-S)}$ denotes the $k$th centroid for channel~$c$.

Given a per-channel target waveform $\mathbf{x}^{(c)}_{\text{tgt}} \in \mathbb{R}^{(W-S)}$, its token target $y^{(c)}$ is obtained by assigning it to the nearest centroid:
\begin{equation}
y^{(c)}
=
\underset{k \in \{1,\ldots,K\}}{\arg\min}\;
\left\|
\mathbf{x}^{(c)}_{\text{tgt}}
-
\boldsymbol{\mu}^{(c)}_{k}
\right\|_{2}
\end{equation}
As a result, the future signal pattern of a multi-sensor window is represented by the channel-wise token vector
\begin{equation}
\mathbf{y}
=
\big(
y^{(1)},\,
y^{(2)},\,
\ldots,\,
y^{(C)}
\big)
\end{equation}
which provides the self-supervision targets used by the prediction head during training.

\subsection{Training Objective}
\label{subsec:objective}

For channel $c$, the training objective is defined as the cross-entropy between the predicted distribution and the ground-truth token class:
\begin{equation}
\label{eq:ce_loss}
\mathcal{L}^{(c)}
=
-\log\!\left(
    \pi^{(c)}_{\, y^{(c)}}
\right)
\end{equation}
where $\mathcal{L}^{(c)}$ is the channel-wise cross-entropy term, and $\pi^{(c)}_{\, y^{(c)}}$ denotes the predicted probability assigned to the ground-truth token class $y^{(c)}$ for channel $c$.

The overall objective for one window is defined as the average of the channel-wise terms:
\begin{equation}
\label{eq:loss}
\mathcal{L}
=
\frac{1}{C}
\sum_{c=1}^{C}
\mathcal{L}^{(c)}
\end{equation}
This objective promotes accurate token prediction and helps the PLM backbone learn temporal and cross-channel patterns in a self-supervised manner.

Fig.\ref{fig:online_training_flow} shows the overall architecture and learning mechanism of LoRM, and Algorithm~\ref{alg:lorm} summarises the LoRM workflow.




\begin{algorithm}[t]
    \caption{Algorithmic Overview of the LoRM Framework}
    \label{alg:lorm}
    \begin{algorithmic}[1]
        \REQUIRE Multi-sensor window $\mathbf{X} \in \mathbb{R}^{W \times C}$; context-segment length $S$; patch length $h$; embedding dimension $d$; per-channel codebooks $\{\mathcal{C}^{(c)}\}_{c=1}^{C}$; pre-trained PLM.
        \ENSURE Updated trainable parameters of LoRM.

        \STATE Initialise trainable parameters:
        embedding matrix $\mathbf{W}_{E}$, positional embedding $\mathbf{P}_{\mathrm{pos}}$, classification matrix $\mathbf{W}_{c}$,
        and the normalisation layers of the PLM.

        \FOR{each window $\mathbf{X}$ in the training set}

            \STATE \textbf{Context-target split}:
            obtain context segment $\mathbf{X}_{\mathrm{ctx}}$ and target segment $\mathbf{X}_{\text{tgt}}$.

            \STATE \textbf{Patching and flattening}:
            divide each channel context segment into patches using~\eqref{eq:patch_def}, and construct the MCPS $\mathbf{P}$ using~\eqref{eq:patch_seq}.

            \STATE \textbf{Patch embedding}:
            compute $\mathbf{E} = \mathbf{P}\mathbf{W}_{E}$ using~\eqref{eq:embed}, then form $\tilde{\mathbf{E}} = \mathbf{E} + \mathbf{P}_{\mathrm{pos}}$ using~\eqref{eq:add_pos}.

            \STATE \textbf{PLM encoding}:
            obtain $\mathbf{Z} = \mathrm{PLM}(\tilde{\mathbf{E}})$ using~\eqref{eq:plm}.

            \STATE \textbf{Global representation}:
            compute $\mathbf{g} = \frac{1}{NC}\,\mathbf{1}^{\top}\mathbf{Z}$ using~\eqref{eq:pool}.

            \STATE \textbf{Token prediction}:
            compute $\mathbf{u} = \mathrm{LayerNorm}(\mathrm{GELU}(\mathbf{g}))$ and $\mathbf{v} = \mathbf{u}\mathbf{W}_{c}$ using~\eqref{eq:head},
            then obtain token probabilities $\pi^{(c)}$ using~\eqref{eq:softmax}.

            \STATE \textbf{Token generation}:
            for each channel $c$, assign the target waveform $\mathbf{x}^{(c)}_{\text{tgt}}$ to the nearest centroid in $\mathcal{C}^{(c)}$ to obtain token $y^{(c)}$.

            \STATE \textbf{Objective computation}:
            compute the channel-wise objective $\mathcal{L}^{(c)}$ using~\eqref{eq:ce_loss}, and the window-level objective $\mathcal{L}$ using~\eqref{eq:loss}.
            
            \STATE  \textbf{Parameters update}:
            Update Trainable parameters using backpropagation and Adam optimiser.

           \ENDFOR
    \end{algorithmic}
\end{algorithm}

\section{LoRM-based Condition Monitoring}
\label{sec:cm}

This section describes the online deployment of LoRM for condition monitoring. After the offline training stage, the trained model is deployed on streaming multi-sensor data for real-time monitoring.

\subsection{Offline Model Training}
\label{Offline_training}

\subsubsection{Step 1 -- Data split and normalisation}
\label{norm}
Given a raw multivariate time series $\mathbf{T}^{\mathrm{raw}}\in\mathbb{R}^{T\times C}$, the data are divided into training and validation subsets, denoted by $\mathbf{T}_{\mathrm{train}}^{\mathrm{raw}}$ and $\mathbf{T}_{\mathrm{val}}^{\mathrm{raw}}$, respectively. For each channel $c$, the mean and standard deviation are computed from the training subset as
\begin{equation}
\label{eq:cm_meanstd}
m^{(c)}=\mathrm{mean}\!\left(\mathbf{T}_{\mathrm{train}}^{\mathrm{raw},(c)}\right),
\qquad
s^{(c)}=\mathrm{std}\!\left(\mathbf{T}_{\mathrm{train}}^{\mathrm{raw},(c)}\right).
\end{equation}
Let $\mathbf{m}=(m^{(1)},\ldots,m^{(C)})$ and $\mathbf{s}=(s^{(1)},\ldots,s^{(C)})$. The training and validation subsets are then normalised channel-wise using the training statistics:
\begin{equation}
\label{eq:cm_norm}
\mathbf{T}_{*}
=
\left(\mathbf{T}_{*}^{\mathrm{raw}}-\mathbf{1}\mathbf{m}^{\top}\right)
\oslash
\left(\mathbf{1}\mathbf{s}^{\top}+\epsilon\right),
\qquad
* \in \{\mathrm{train},\mathrm{val}\}
\end{equation}
where $\oslash$ denotes element-wise division and $\epsilon$ is a small constant for numerical stability.

\subsubsection{Step 2 -- Self-supervised training of LoRM}
The training procedure of LoRM is summarised as follows.
\begin{itemize}
    \item \textit{Windowing:} The normalised training dataset $\mathbf{T}_{\mathrm{train}}$ is segmented into $N_{\mathrm{tr}}$ windows $\{\mathbf{X}^{\mathrm{train}}_{k_1}\in\mathbb{R}^{W\times C}\}_{k_1=1}^{N_{\mathrm{tr}}}$. Similarly, the validation dataset $\mathbf{T}_{\mathrm{val}}$ is segmented into $N_{\mathrm{val}}$ windows $\{\mathbf{X}^{\mathrm{val}}_{k_2}\in\mathbb{R}^{W\times C}\}_{k_2=1}^{N_{\mathrm{val}}}$, where $k_1$ and $k_2$ denote the window indices.

    \item \textit{Model training:} The training windows $\mathbf{X}^{\mathrm{train}}_{k_1}$ are used to compute the training objective in Algorithm~\ref{alg:lorm}. Model parameters are updated using the Adam optimiser \cite{ADAM}. Only a lightweight set of parameters is optimised, including $\mathbf{W}_{E}$, $\mathbf{P}_{\mathrm{pos}}$, LayerNorm parameters, and the classification head.

    \item \textit{Model validation:} After each training epoch, the current model is evaluated on $\mathbf{X}^{\mathrm{val}}_{k_2}$ to obtain the validation objective. The checkpoint achieving the minimum validation objective is saved.

    \item \textit{Model selection:} Training is terminated by early stopping when the validation objective does not improve for a sustained period. The checkpoint with the minimum validation objective is retained as the trained LoRM model, denoted by $\mathcal{M}$.
\end{itemize}

\subsection{Online Condition Monitoring}

\subsubsection{Step 1 -- Data normalisation}
After the offline stage, the trained LoRM model $\mathcal{M}$ is deployed to continuously evaluate the incoming multi-sensor data stream for condition monitoring. The streaming signal is segmented into consecutive windows $\{\hat{\mathbf{X}}^{\mathrm{ol}}_{k_3}\in\mathbb{R}^{W\times C}\}_{k_3=1}^{\infty}$, where $k_3$ denotes the window index. Each window is normalised channel-wise as
\begin{equation}
\label{eq:online_window}
\mathbf{X}^{\mathrm{ol}}_{k_3}
=
\left(\hat{\mathbf{X}}^{\mathrm{ol}}_{k_3}-\mathbf{1}_{W}\mathbf{m}^{\top}\right)
\oslash
\left(\mathbf{1}_{W}\mathbf{s}^{\top}+\epsilon\right)
\end{equation}
where $\mathbf{1}_{W}\in\mathbb{R}^{W\times 1}$ is an all-ones vector and $\epsilon$ is a small constant for numerical stability. The normalised window is then divided into a context segment and a target segment, where the context segment is used as the model input.

\subsubsection{Step 2 -- Condition monitoring}
As the system health condition changes, the distribution of the multi-sensor data deviates from the normal evolution patterns learned by the model, leading to degraded token prediction performance. Therefore, the model prediction error can be used as a health indicator to infer trends in the system state.


For each incoming window $\mathbf{X}^{\mathrm{ol}}_{k_3}$, a window-level prediction score is computed as
\begin{equation}
\label{eq:rt_wfl}
\mathrm{WLF}(\mathbf{X}^{\mathrm{ol}}_{k_3})
=
\mathcal{L}(\mathcal{M}(\mathbf{X}^{\mathrm{ol}}_{k_3}))
\end{equation}
where $\mathrm{WLF}(\cdot)$ denotes the window-level prediction score, measuring the mismatch between the predicted token distribution and the corresponding target tokens for that window.

The health index, denoted by $\mathrm{HI}$, is defined as the deviation of the current score from an initial online baseline computed over the first $\Delta$ windows:
\begin{equation}
\label{eq:rt_hi}
\mathrm{HI}(\mathbf{X}^{\mathrm{ol}}_{k_3})
=
\mathrm{WLF}(\mathbf{X}^{\mathrm{ol}}_{k_3})
-
\frac{1}{\Delta}\sum_{\delta=1}^{\Delta}\mathrm{WLF}(\mathbf{X}^{\mathrm{ol}}_{\delta})
\end{equation}
where $\Delta$ denotes the initial buffer length used to estimate the baseline. An alarm is triggered when
\begin{equation}
\label{eq:rt_threshold}
\mathrm{HI}(\mathbf{X}^{\mathrm{ol}}_{k_3}) > \tau
\end{equation}
where $\tau$ denotes a predefined monitoring threshold.

\begin{figure*}[t]
    \centering
    \includegraphics[width=0.90\textwidth]{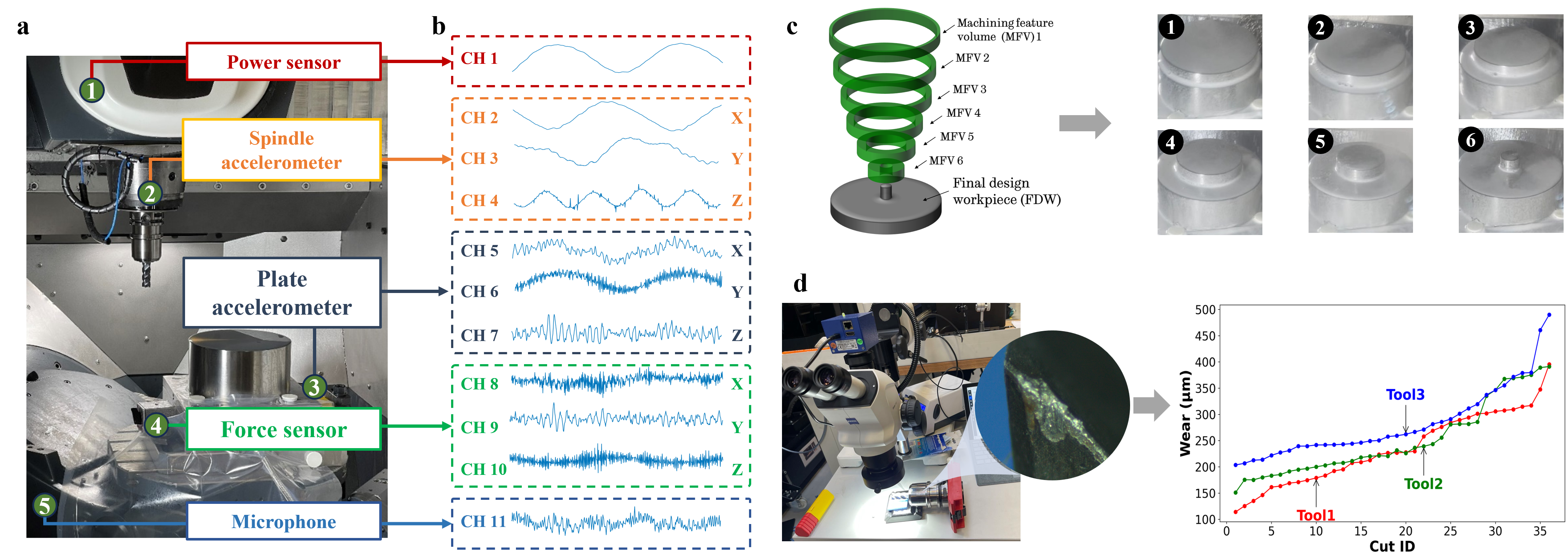}
    \caption{{\color{black}Overview of the TCM case study. (a) Front view of the DMU 40 eVo CNC milling machine and the deployed sensing system, including a power sensor, tri-axis spindle accelerometer, a tri-axis plate accelerometer, a tri-axis force sensor and a microphone. (b) Examples of the collected multi-modal sensor signals. (c) Illustration of the dynamic milling strategy. (d) Evolution of the maximum tool wear measured from the four-flute cutters for Tool 1 to 3 using a microscope.}}
    \label{Overall_Experiment}
\end{figure*}

\subsection{Parameter Discussion}
The implementation of LoRM involves several hyperparameters, including the context-segment length $S$, patch length $h$, number of clusters $K$, embedding dimension $d$, initial buffer length $\Delta$, and monitoring threshold $\tau$. In general, $S$ and $h$ control the trade-off between temporal coverage and computational cost, $K$ determines the granularity of discrete token states, and $d$ is determined by the hidden size of the chosen PLM backbone. For online monitoring, $\Delta$ controls the stability of the initial baseline estimation, while $\tau$ determines the sensitivity of alarm triggering. In this study, these parameters were selected empirically to balance predictive performance, computational efficiency, and monitoring stability.

\section{Field Experiment}
\subsection{Design of Experiment}
To validate the proposed LoRM framework for rotating machinery condition monitoring, a real-world tool condition monitoring (TCM) case study was conducted at the Advanced Manufacturing Research Centre (AMRC), University of Sheffield, UK, as shown in Fig.~\ref{Overall_Experiment}(a). Run-to-failure tests were carried out on a 5-axis DMU 40 eVo CNC milling machine to obtain full life-cycle tool wear data. The cutting tools were Sandvik 4-flute carbide end mills, and the workpieces were made of Ti-6Al-4V~\cite{LIU2024163}.

In terms of the machining parameters, the spindle speed was 2586~r/min. The feed rate and table feed were 0.092~mm/tooth and 955.04~mm/min, respectively. The axial depth of cut (ADoC), radial depth of cut (RDoC), and tool radius were 19.5\%, 12\%, and 0.06~mm, respectively.

To track the cutting dynamics and tool degradation, a multi-modal sensing system was deployed, as shown in Fig.~\ref{Overall_Experiment}(a). The system was used to record vibration, power, acoustic, and force signals using the following sensors:
\begin{itemize}
    \item Spindle accelerometer (3-axis), PCB type, sensitivity 10 mV/g, bandwidth 1--5 kHz, measuring spindle vibration signals in the X, Y, and Z directions
    \item Plate accelerometer (3-axis), PCB type, sensitivity 500 mV/g, bandwidth 1--5 kHz, measuring plate vibration signals in the X, Y, and Z directions
    \item Force sensor (3-axis), KISTLER type, sensitivity -7.8 pC/N, measuring cutting force in the X, Y, and Z directions
    \item Power sensor for collecting spindle power signals
    \item Microphone for capturing acoustic signals
\end{itemize}
As shown in Fig.~\ref{Overall_Experiment}(b), the sensing system provided 11 measurement channels, comprising three directions (X, Y, and Z) from the spindle accelerometer, three directions (X, Y, and Z) from the plate accelerometer, three directions (X, Y, and Z) from the force sensor, one power channel, and one microphone channel. In this study, the sampling rate of the sensing system was set to 51.2 kHz.

During the machining trials, three cutting tools, denoted as T1, T2, and T3, were used to machine the workpieces. A dynamic milling strategy was adopted, in which each workpiece was processed in four layers. Each layer comprised six ring-shaped machining feature volumes (MFVs), as illustrated in Fig.~\ref{Overall_Experiment}(c). After completing the final ring cut of a layer, the workpiece surface was cleaned before proceeding to the next set of ring cuts. Following each ring cut, the machine was paused and tool wear was inspected using a microscope (Fig.~\ref{Overall_Experiment}(d)). By the end of the experiments, each tool had machined six layers (one and a half workpieces), resulting in 36 ring cuts in total per tool. Fig.~\ref{Overall_Experiment}(d) presents the evolution of the maximum flank wear measured on the 4-flute cutters. The most severe wear increased from approximately 150~$\mu$m to above 400~$\mu$m.

\subsection{Offline Model Training}
\label{subsec:offline_training}

The data collected from T1 were first used for offline training. Following Section~\ref{Offline_training}, the offline stage consists of data split, data normalisation, and self-supervised training of LoRM. Specifically, the data from the first six ring cuts of T1 were used for model training. Model performance was then evaluated in three separate tests: (i) the remaining ring cuts of T1 to assess within-tool generalisation and determine the monitoring threshold, (ii) the complete dataset from T2, and (iii) the complete dataset from T3 to assess cross-tool generalisation.

The data from the first six ring cuts were further randomly divided into a training subset \(\mathbf{T}_{\mathrm{train}}\) (80\%) and a validation subset \(\mathbf{T}_{\mathrm{val}}\) (20\%). After normalisation, the training subset was segmented into fixed-length windows, where each window contained 321 samples per channel. In this study, each window was divided into a continuous context segment containing 320 samples and a target segment containing 1 sample. The context segment was further converted into the MCPS through patching and flattening, and the resulting MCPS served as the model input. For self-supervision, the target segment of each channel was quantised into a discrete token target via per-channel \(k\)-means clustering. The channel-wise codebooks were constructed using target-segment samples from \(\mathbf{T}_{\mathrm{train}}\). In this study, each channel was quantised into \(K = 10\) clusters, meaning that 10 token classes were generated.

For model training, we used pre-trained GPT-2 as the backbone of LoRM \cite{Radford2019LanguageMA}. The experiments were implemented in Python 3.10.12 on an Ubuntu system with 24 GB memory and an NVIDIA GeForce RTX 4090 GPU. Only lightweight components, including the embedding matrix, positional embedding, LayerNorm parameters, and prediction head, were updated during training, while the backbone attention and feed-forward blocks were frozen. Validation objective was monitored for early stopping, and the best-performing checkpoint was retained as the final LoRM model.


\begin{table}[ht]
    \centering
    \caption{Classification Performance Metrics}
    \label{tab:metrics}
    \begin{threeparttable}
    \setlength{\tabcolsep}{4pt}
    \begin{tabular}{@{} l c >{\centering\arraybackslash}p{4cm} @{}}
        \toprule
        \textbf{Metric} & \textbf{Notation} & \textbf{Formula} \\
        \midrule
        Accuracy & ACC & $\frac{TP + TN}{TP + TN + FP + FN}$ \\
        \addlinespace[2pt]
        Precision & P & $\frac{TP}{TP + FP}$ \\
        \addlinespace[2pt]
        Recall & R & $\frac{TP}{TP + FN}$ \\
        \addlinespace[2pt]
        F1-score & F1 & $2 \times \frac{P \times R}{P + R}$ \\
        \addlinespace[2pt]
        False positive rate & FPR & $\frac{FP}{FP + TN}$ \\
        \bottomrule
    \end{tabular}
    \begin{tablenotes}[flushleft]
        \footnotesize
        \item Here, the positive class denotes abnormal windows and the negative class denotes healthy windows; TP = true positive, TN = true negative, FP = false positive, and FN = false negative.
    \end{tablenotes}
    \end{threeparttable}
\end{table}

\begin{table*}
\centering
\scriptsize
\setlength{\tabcolsep}{2.2pt}
\renewcommand{\arraystretch}{1}
\caption{Cross-Tool Prediction Performance of Different Methods}
\label{Whole_Results}
\resizebox{\textwidth}{!}{%
\begin{tabular}{lllcccccccccccccccccc}
\toprule
Method & Backbone & Test Tool
& \multicolumn{6}{c}{Train on Tool1}
& \multicolumn{6}{c}{Train on Tool2}
& \multicolumn{6}{c}{Train on Tool3} \\
\cmidrule(lr){4-9}\cmidrule(lr){10-15}\cmidrule(lr){16-21}
& & &
Acc & P & R & F1 & FPR & \shortstack{Error$\downarrow$\\($\mu$m)} &
Acc & P & R & F1 & FPR & \shortstack{Error$\downarrow$\\($\mu$m)} &
Acc & P & R & F1 & FPR & \shortstack{Error$\downarrow$\\($\mu$m)} \\
\midrule
\multirow{4}{*}{\shortstack{Traditional}}
& \multirow{4}{*}{Statistics\cite{CAO2008141}}
& Tool1 & 0.98 & 1.00 & 0.91 & 0.95 & 0.00 & 1.21 & 0.96 & 1.00 & 0.79 & 0.88 & 0.00 & 5.62 & 0.79 & 0.00 & 0.00 & 0.00 & 0.01 & 185.95 \\
& & Tool2 & 0.92 & 1.00 & 0.54 & 0.71 & 0.00 & 79.00 & 0.93 & 1.00 & 0.60 & 0.75 & 0.00 & 35.18 & 0.87 & 1.00 & 0.31 & 0.48 & 0.00 & 70.96 \\
& & Tool3 & 0.88 & 1.00 & 0.53 & 0.70 & 0.00 & 66.23 & 0.98 & 1.00 & 0.94 & 0.97 & 0.00 & 11.26 & 0.99 & 1.00 & 0.95 & 0.97 & 0.00 & 1.15 \\
& & Average & 0.93 & 1.00 & 0.66 & 0.79 & 0.00 & 48.81 & 0.96 & 1.00 & 0.78 & 0.87 & 0.00 & 17.35 & 0.88 & 0.67 & 0.42 & 0.48 & 0.00 & 86.02 \\
\midrule
\multirow{16}{*}{\shortstack{PLM\\-based}}
& \multirow{4}{*}{\shortstack{GPT2\\(Pretrained)}\cite{Radford2019LanguageMA}}
& Tool1 & 0.90 & 0.68 & 1.00 & 0.81 & 0.13 & 23.99 & 0.97 & 1.00 & 0.86 & 0.92 & 0.00 & 1.96 & 0.84 & 1.00 & 0.27 & 0.42 & 0.00 & 7.51 \\
& & Tool2 & 0.98 & 1.00 & 0.89 & 0.94 & 0.00 & 35.18 & 0.99 & 0.99 & 0.95 & 0.97 & 0.00 & 35.18 & 0.93 & 0.76 & 0.96 & 0.85 & 0.08 & 18.42 \\
& & Tool3 & 0.93 & 0.82 & 0.97 & 0.89 & 0.09 & 14.39 & 0.79 & 1.00 & 0.25 & 0.40 & 0.00 & 11.26 & 0.98 & 0.94 & 0.99 & 0.96 & 0.03 & 1.15 \\
& & Average & 0.94 & 0.83 & 0.95 & 0.88 & 0.07 & 24.52 & 0.91 & 1.00 & 0.68 & 0.76 & 0.00 & 16.13 & 0.91 & 0.90 & 0.74 & 0.74 & 0.03 & 9.03 \\
\cmidrule(lr){2-21}

& \multirow{4}{*}{BERT\cite{Bert}}
& Tool1 & 0.91 & 0.72 & 1.00 & 0.84 & 0.11 & 15.65 & 0.77 & 0.49 & 0.83 & 0.62 & 0.24 & 153.88 & 0.78 & 0.00 & 0.00 & 0.00 & 0.00 & \multicolumn{1}{c}{\textit{N/A}} \\
& & Tool2 & 0.96 & 1.00 & 0.82 & 0.90 & 0.00 & 46.31 & 0.99 & 0.99 & 0.94 & 0.96 & 0.00 & 35.18 & 0.79 & 0.00 & 0.00 & 0.00 & 0.00 & \multicolumn{1}{c}{\textit{N/A}} \\
& & Tool3 & 0.87 & 0.71 & 0.92 & 0.80 & 0.15 & 55.75 & 0.92 & 0.90 & 0.80 & 0.85 & 0.03 & 86.42 & 0.96 & 0.90 & 0.97 & 0.93 & 0.04 & 9.21 \\
& & Average & 0.92 & 0.81 & 0.92 & 0.85 & 0.09 & 39.24 & 0.89 & 0.79 & 0.86 & 0.81 & 0.09 & 91.83 & 0.84 & 0.30 & 0.32 & 0.31 & 0.01 & \multicolumn{1}{c}{\textit{N/A}} \\
\cmidrule(lr){2-21}

& \multirow{4}{*}{T5\cite{T5}}
& Tool1 & 0.91 & 0.72 & 1.00 & 0.84 & 0.11 & 15.65 & 0.75 & 0.46 & 0.69 & 0.55 & 0.24 & 153.88 & 0.78 & 0.00 & 0.00 & 0.00 & 0.00 & \multicolumn{1}{c}{\textit{N/A}} \\
& & Tool2 & 0.96 & 1.00 & 0.80 & 0.89 & 0.00 & 67.55 & 0.99 & 1.00 & 0.97 & 0.98 & 0.00 & 35.18 & 0.79 & 0.00 & 0.00 & 0.00 & 0.00 & \multicolumn{1}{c}{\textit{N/A}} \\
& & Tool3 & 0.93 & 0.91 & 0.81 & 0.86 & 0.03 & 9.21 & 0.95 & 0.84 & 1.00 & 0.91 & 0.07 & 14.39 & 0.94 & 0.93 & 0.86 & 0.89 & 0.03 & 1.15 \\
& & Average & 0.93 & 0.88 & 0.87 & 0.86 & 0.05 & 30.80 & 0.90 & 0.77 & 0.89 & 0.82 & 0.10 & 67.82 & 0.84 & 0.31 & 0.29 & 0.30 & 0.01 & \multicolumn{1}{c}{\textit{N/A}} \\
\cmidrule(lr){2-21}

& \multirow{4}{*}{\shortstack{GPT2\\(Scratch)}\cite{Radford2019LanguageMA}}
& Tool1 & 0.92 & 0.73 & 1.00 & 0.84 & 0.11 & 15.65 & 0.77 & 0.45 & 0.27 & 0.34 & 0.09 & 153.88 & 0.78 & 0.00 & 0.00 & 0.00 & 0.00 & \multicolumn{1}{c}{\textit{N/A}} \\
& & Tool2 & 0.96 & 1.00 & 0.79 & 0.89 & 0.00 & 67.55 & 0.99 & 1.00 & 0.93 & 0.96 & 0.00 & 35.18 & 0.79 & 0.00 & 0.00 & 0.00 & 0.00 & \multicolumn{1}{c}{\textit{N/A}} \\
& & Tool3 & 0.72 & 0.00 & 0.00 & 0.00 & 0.00 & \multicolumn{1}{c}{\textit{N/A}} & 0.72 & 0.00 & 0.00 & 0.00 & 0.00 & \multicolumn{1}{c}{\textit{N/A}} & 0.98 & 0.94 & 0.99 & 0.96 & 0.03 & 1.15 \\
& & Average & 0.86 & 0.58 & 0.60 & 0.58 & 0.04 & \multicolumn{1}{c}{\textit{N/A}} & 0.82 & 0.48 & 0.40 & 0.44 & 0.03 & \multicolumn{1}{c}{\textit{N/A}} & 0.85 & 0.31 & 0.33 & 0.32 & 0.01 & \multicolumn{1}{c}{\textit{N/A}} \\
\midrule

\multirow{4}{*}{\shortstack{PSM\\-based}}
& \multirow{4}{*}{Whisper\cite{pmlr-v202-radford23a}}
& Tool1 & 0.90 & 0.68 & 1.00 & 0.81 & 0.13 & 23.99 & 0.99 & 0.98 & 0.97 & 0.97 & 0.01 & 5.93 & 0.86 & 1.00 & 0.37 & 0.54 & 0.00 & 14.78 \\
& & Tool2 & 0.96 & 1.00 & 0.80 & 0.89 & 0.00 & 67.55 & 0.99 & 0.99 & 0.94 & 0.96 & 0.00 & 35.18 & 0.87 & 1.00 & 0.39 & 0.56 & 0.00 & 70.96 \\
& & Tool3 & 0.81 & 1.00 & 0.34 & 0.50 & 0.00 & 46.23 & 0.97 & 0.93 & 0.95 & 0.94 & 0.03 & 9.21 & 0.95 & 0.89 & 0.94 & 0.91 & 0.05 & 9.21 \\
& & Average & 0.89 & 0.89 & 0.71 & 0.73 & 0.04 & 45.92 & 0.98 & 0.97 & 0.95 & 0.96 & 0.01 & 16.77 & 0.89 & 0.96 & 0.57 & 0.67 & 0.02 & 31.65 \\
\midrule

\multirow{4}{*}{\shortstack{Time\\-series}}
& \multirow{4}{*}{GPT4TS\cite{zhou2023onefitsall}}
& Tool1 & 0.78 & 0.50 & 1.00 & 0.67 & 0.28 & 72.36 & 0.51 & 0.31 & 1.00 & 0.47 & 0.63 & 165.03 & 0.57 & 0.17 & 0.23 & 0.19 & 0.33 & 174.62 \\
& & Tool2 & 0.98 & 1.00 & 0.90 & 0.95 & 0.00 & 46.31 & 1.00 & 1.00 & 1.00 & 1.00 & 0.00 & 35.18 & 0.95 & 1.00 & 0.78 & 0.87 & 0.00 & 67.55 \\
& & Tool3 & 0.77 & 0.55 & 1.00 & 0.71 & 0.31 & 42.33 & 0.77 & 0.55 & 1.00 & 0.71 & 0.31 & 42.33 & 0.95 & 0.87 & 0.97 & 0.92 & 0.06 & 9.21 \\
& & Average & 0.85 & 0.69 & 0.97 & 0.78 & 0.20 & 53.67 & 0.76 & 0.62 & 1.00 & 0.73 & 0.32 & 80.85 & 0.83 & 0.68 & 0.66 & 0.66 & 0.13 & 83.79 \\

\bottomrule
\end{tabular}%
}
\end{table*}

\subsection{Online Condition Monitoring}
After offline training, the trained LoRM model was deployed for real-time condition monitoring on streaming multi-sensor signals.

\begin{figure}[tb]
	\centering
	\subfigure[]{
		\label{all_tests_merged_a}
		\includegraphics[width=0.98\columnwidth]{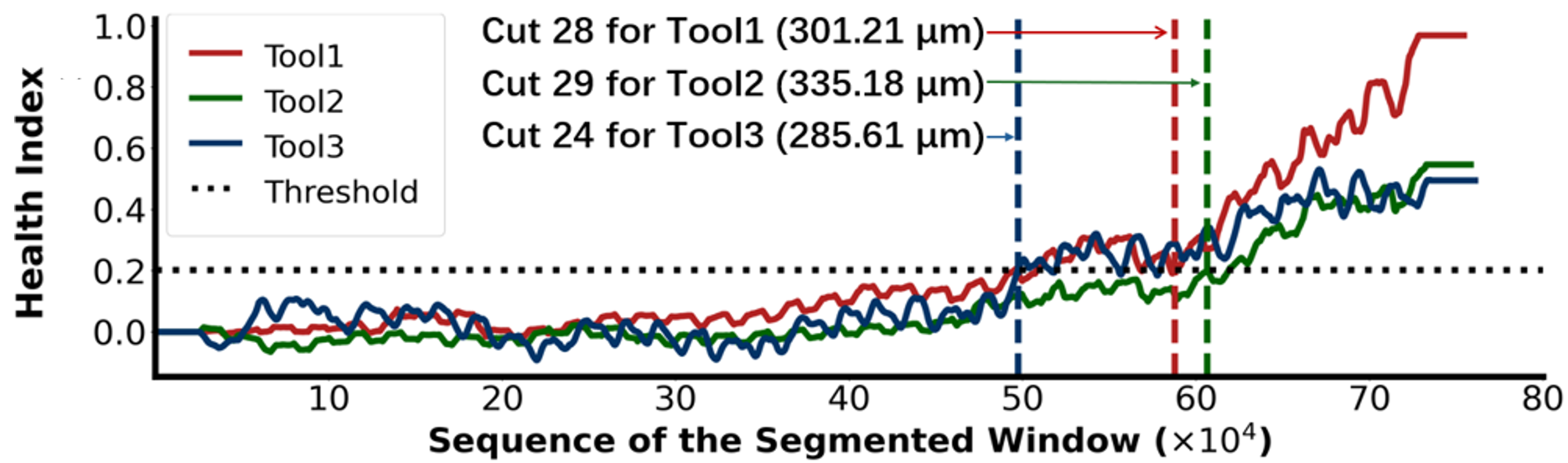}}
	\subfigure[]{
		\label{all_tests_merged_b}
		\includegraphics[width=0.98\columnwidth]{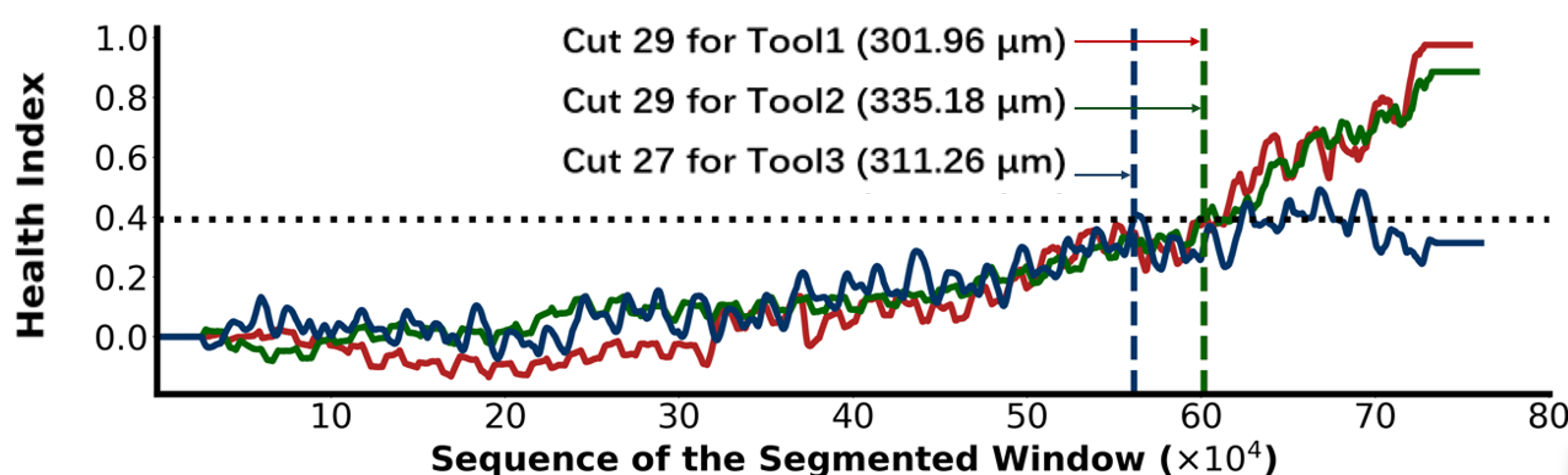}}
    \subfigure[]{
		\label{all_tests_merged_c}
		\includegraphics[width=0.98\columnwidth]{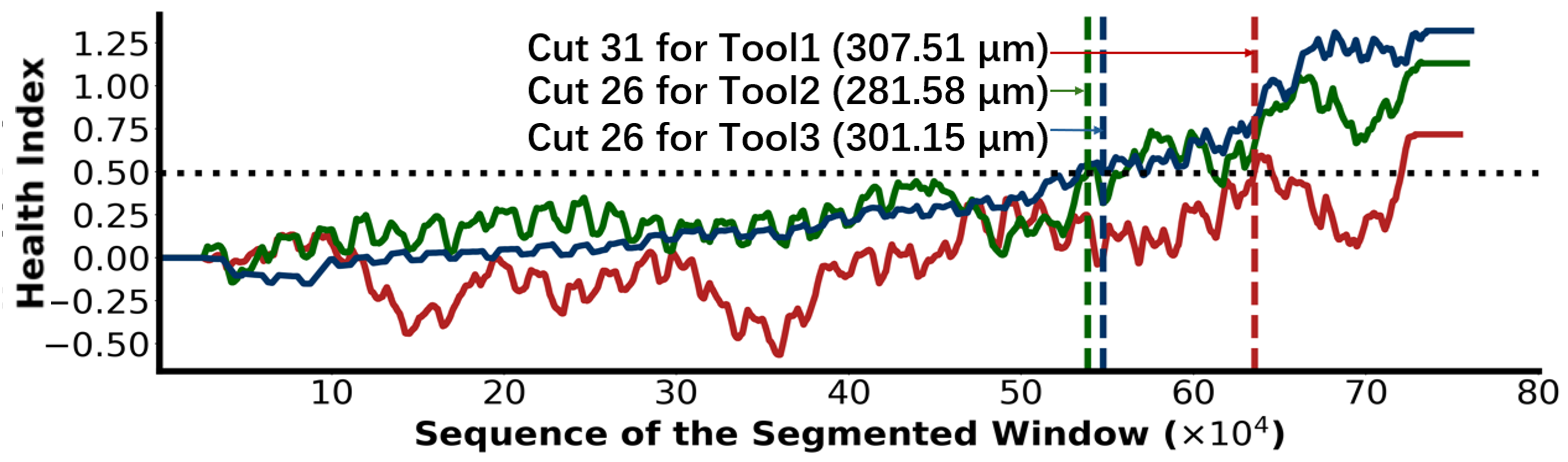}}    
    \caption{Field-test results of \textsc{LoRM} under cross-tool settings: (a) trained on T1 and tested on T1, T2, and T3; (b) trained on T2 and tested on T1, T2, and T3; and (c) trained on T3 and tested on T1, T2, and T3.}
	\label{all_tests_merged}
\end{figure}

We first evaluated the monitoring behavior on T1. To avoid information leakage, the LoRM model was trained only on the first six ring cuts of T1, while all online monitoring results reported for T1 were obtained from the remaining ring cuts of T1. For each window, LoRM outputs a token probability distribution for each channel, from which the WLF is computed. Based on the WLF, the HI was computed, where the initial buffer length \(\Delta\) was set to 20000. As shown in Fig.~\ref{all_tests_merged_a}, the HI exhibits an overall increasing trend. Under healthy operation, the target tokens follow the stable dynamics learned from the training data and can be predicted accurately, resulting in a low WLF. As wear progresses, the signal distribution drifts, leading to larger prediction errors and hence increased WLF and HI.

Next, we calibrated the monitoring threshold using the HI curve of T1. According to the ISO 3685:1993 standard \cite{ISO3685}, the tool-wear threshold is \(300~\mu\mathrm{m}\). Based on the measured tool wear in Fig.~\ref{Overall_Experiment}(d), tool wear was recorded only at discrete ring cuts. Therefore, we selected Cut~28, whose measured wear (\(301.21~\mu\mathrm{m}\)) is the closest to the ISO threshold of \(300~\mu\mathrm{m}\), to calibrate the monitoring threshold. As shown in Fig.~\ref{all_tests_merged_a}, the corresponding HI value at Cut~28 is 0.20. This value was therefore selected as the monitoring threshold \(\tau\), which was then kept fixed for subsequent cross-tool evaluation.

Finally, we tested the trained model on T2 and T3 using the same threshold \(\tau\). As shown in Fig.~\ref{all_tests_merged_a}, for T2, the HI first exceeds \(\tau\) at Cut~29, where the measured tool wear is \(335.18~\mu\mathrm{m}\), corresponding to a deviation of \(35.18~\mu\mathrm{m}\) from the \(300~\mu\mathrm{m}\) wear threshold. For T3, the HI first exceeds \(\tau\) at Cut~24, where the measured tool wear is \(285.61~\mu\mathrm{m}\), corresponding to a deviation of \(14.39~\mu\mathrm{m}\).



Finally, we further evaluated performance using the standard metrics summarised in Table~\ref{tab:metrics}, including Accuracy (ACC), Precision (P), Recall (R), F1-score, and False Positive Rate (FPR). For online condition monitoring, Precision and Recall are important because they reflect false alarms and missed detections, respectively. To derive window-level ground truth, the measured tool wear for each ring cut was assigned to all windows within that cut, and each window was labelled as abnormal if its wear exceeded \(300~\mu\mathrm{m}\) and healthy otherwise, with abnormal and healthy windows treated as the positive and negative classes, respectively. The results are summarised in Table~\ref{Whole_Results}.

\subsection{Additional cross-tool validation}
To further examine the generality of LoRM, we repeated the evaluation by using each tool in turn as the development tool. Specifically, we repeated the same offline training and online monitoring procedure with T2 as the development tool (training/validation on the first six ring cuts of T2, with the remaining ring cuts of T2 used for threshold calibration), and then evaluated the resulting model on T1 and T3. Similarly, we trained LoRM using T3 as the development tool and evaluated it on T1 and T2. In all settings, the monitoring threshold \(\tau\) was determined using the corresponding development tool only and then kept fixed when testing on the other tools. The results are presented in Figs.~\ref{all_tests_merged_b} and \ref{all_tests_merged_c} and summarised in Table~\ref{Whole_Results}. As can be seen, the two repeated settings show comparable performance. In particular, the average prediction errors of the models developed using T2 and T3 are \(16.13~\mu\mathrm{m}\) and \(9.03~\mu\mathrm{m}\), respectively. These errors are close to that of the model developed using T1 (24.52~\(\mu\mathrm{m}\)), indicating that LoRM remains robust under different tool initialisations.

\section{Comparative Study}
\subsection{Comparison with Traditional Statistical Feature Methods}

We first compared \textsc{LoRM} with a conventional statistical-feature baseline. As shown in Table~\ref{Whole_Results}, this baseline is generally more conservative, tending to avoid predicting abnormal states. It leads to a very low false positive rate (FPR) across tool conditions which is desirable in itself, while its Recall is also substantially lower. Specifically, when trained on T1, T2, and T3, its average Recall is only 0.66, 0.78, and 0.42, respectively. This indicates that the model is strongly biased towards classifying tool states as normal, resulting in frequent missed detections under fault conditions.

Although the traditional baseline also achieves relatively high Accuracy in some settings, this does not necessarily reflect reliable monitoring performance. During testing, healthy windows are much more frequent than abnormal ones, so a conservative model can obtain high Accuracy simply by favouring the majority normal class. In this case, the apparently high Accuracy is largely driven by class imbalance rather than by genuinely effective state discrimination. Such behaviour is undesirable for condition monitoring, because a method that rarely raises false alarms but frequently misses true degradation cannot provide reliable early warning. In addition, the statistical baseline produces larger absolute wear errors in several settings, further indicating weaker cross-tool robustness than \textsc{LoRM}.

\subsection{Backbone Comparison}
To examine the sensitivity of \textsc{LoRM} to the PLM backbone, we kept the pipeline unchanged and varied only the backbone. All other settings were identical.

\subsubsection{Effect of Architecture: Pretrained Decoder-Only vs.\ Encoder-Only Backbones}
We compared a pretrained decoder-only backbone (GPT-2) with pretrained encoder-only language models (BERT and T5). As shown in Table~\ref{Whole_Results}, when trained on T1 or T2, BERT and T5 sometimes achieved accuracy comparable to GPT-2, but they generally produced larger absolute wear errors. For example, under the Train-on-T2 setting, pretrained GPT-2 achieved an average error of only $16.13~\mu$m, compared with $91.83~\mu$m for BERT and $67.82~\mu$m for T5. This difference became more pronounced under the Train-on-T3 setting: BERT and T5 performed poorly on both T1 and T2, whereas GPT-2 remained effective across all test tools. This suggests that \textsc{LoRM} is sensitive to backbone architecture under cross-tool conditions. This may be partly attributed to objective alignment, since GPT-2 is pretrained autoregressively, which is closer to our target-token prediction objective than masked language modelling.

\subsubsection{Effect of Pre-Training: Pretrained GPT-2 vs.\ GPT-2 Trained from Scratch}
We replaced the pretrained GPT-2 backbone with the same GPT-2 architecture initialised from scratch and trained end-to-end without freezing. As shown in Table~\ref{Whole_Results}, the scratch model degraded markedly in cross-tool evaluation, whereas the pretrained model remained robust. For example, under the Train-on-T2 setting, pretrained GPT-2 achieved an average F1-score of 0.76, compared with only 0.44 for GPT-2 trained from scratch. Under the Train-on-T3 setting, the pretrained model achieved an average error of $9.03~\mu$m, whereas the scratch model failed in multiple cross-tool cases. These results suggest that large-scale pre-training provides a transferable sequence-learning prior for tokenised signals, thereby improving cross-tool generalisation.

\subsubsection{Effect of Domain: Speech-Pretrained vs.\ Text-Pretrained Backbones}
We used the encoder of a speech-pretrained model (Whisper) as the backbone and compared it with text-pretrained PLMs. As shown in Table~\ref{Whole_Results}, Whisper performed close to text-pretrained GPT-2 and remained effective under cross-tool evaluation. For example, under the Train-on-T2 setting, Whisper achieved an average F1-score of 0.96 and an average error of $16.77~\mu$m, which is comparable to pretrained GPT-2 (F1-score of 0.76 and error of $16.13~\mu$m), although with a different Precision--Recall trade-off. Under the Train-on-T3 setting, Whisper still achieved an average F1-score of 0.67, indicating that \textsc{LoRM} can benefit from both speech and text pre-training.

\subsection{Ablation Study: Discrete vs.\ Continuous Target Prediction}
We compared \textsc{LoRM} with a continuous regression baseline (GPT4TS), which regresses future waveforms rather than predicting discrete token states. Predicting discrete target tokens is better aligned with the original token-based pre-training objective of language models than direct continuous waveform regression. Moreover, compared with GPT4TS and other time-series models, \textsc{LoRM} can capture cross-channel dependencies through its flattening design, whereas conventional time-series models mainly focus on intra-channel dependencies. As shown in Table~\ref{Whole_Results}, GPT4TS yielded substantially larger wear errors and higher FPR under cross-tool conditions. For example, under the Train-on-T2 setting, GPT4TS achieved an average Recall of 1.00 but also a high average FPR of 0.32, together with an average error of $80.85~\mu$m. In contrast, the pretrained GPT-2 backbone used in \textsc{LoRM} achieved a lower Recall of 0.68 but reduced the average FPR to 0.00 and the average error to only $16.13~\mu$m. Similarly, under the Train-on-T3 setting, GPT4TS yielded an average error of $83.79~\mu$m, whereas the pretrained GPT-2 backbone reduced this error to $9.03~\mu$m. These results suggest that the discrete-token formulation of \textsc{LoRM} is more stable and better suited to practical condition monitoring.

\section{Discussion}
This study leads to four main observations:
\begin{itemize}
    \item LoRM is sensitive to the backbone choice in cross-tool evaluation. Autoregressive pre-training is better aligned with target-token prediction, and text pre-training provides a transferable initialisation for tokenised signals, improving cross-tool robustness over training from scratch. This is consistent with the short-window predictability of rotating-machinery signals, where the near past is informative for near-future state prediction.

    \item Pre-training is important for robust transfer. Compared with GPT2 trained from scratch, the pretrained GPT2 backbone achieves more stable cross-tool performance and avoids collapse in several cross-tool settings. This suggests that large-scale pre-training provides a useful sequence-modelling prior for tokenised rotating-machinery signals.

    \item Discrete target-token prediction improves robustness. Compared with continuous forecasting (e.g., GPT4TS), predicting discrete future states is more stable because tokenisation converts waveform regression into classification over a finite codebook, thereby reducing ambiguity, noise sensitivity, and error accumulation.

    \item  LoRM provides a better balance between false alarms and missed detections than the baselines. In contrast, GPT4TS tends to favour Recall at the cost of a higher FPR, while the traditional statistical baseline is overly conservative: although its FPR is very low, its Recall is much lower under cross-tool conditions, meaning that many abnormal windows are missed. Such behaviour is also undesirable in practice, because a monitoring system that rarely raises false alarms but frequently misses real degradation cannot provide reliable early warning.
    

\end{itemize}

\section{Conclusion}
This paper proposes LoRM, a framework that treats multi-modal rotating-machinery signals as a learnable language and reformulates signal understanding as a token-based sequence-prediction problem. By discretising future target segments and predicting target tokens from observed context, LoRM enables self-supervised learning without relying on hand-crafted transforms or features. Condition monitoring is achieved by tracking token-prediction errors as a health indicator. In-situ tool condition monitoring (TCM) experiments show that LoRM achieves stable real-time tracking and strong cross-tool generalisation, outperforming both a continuous waveform prediction baseline, such as GPT4TS, and a traditional statistical-feature method. Future work will focus on extending LoRM to broader rotating-machinery scenarios and richer sensing modalities under varying operating conditions.

\bibliographystyle{Bibliography/IEEEtranTIE}
\bibliography{Bibliography/IEEEabrv,Bibliography/mybibfile}\ 

\end{document}